\documentclass[10pt,twocolumn,letterpaper]{article}

\usepackage{times}
\usepackage{epsfig}
\usepackage{graphicx}
\usepackage{amsmath}
\usepackage{amssymb}
\usepackage{sidecap}
\usepackage{epstopdf}
\usepackage{multirow}

\usepackage{caption}
\usepackage{subcaption}
\usepackage[strings]{underscore}

\newsavebox\CBox

\newcommand{\eg}{{\emph{e.g.},\ }}
\newcommand{\ie}{{\emph{i.e.},\ }}

\usepackage[pagebackref=true,breaklinks=true,letterpaper=true,colorlinks,bookmarks=false]{hyperref}
\usepackage[top = 2.00cm, bottom = 2.00cm, left = 1.95cm, right = 1.95cm]{geometry}

\begin{document}

\title{\fontsize{22}{30}\selectfont AI Benchmark: Running Deep Neural Networks \\ on Android Smartphones \vspace{4mm}}

\author{Andrey Ignatov\\
\footnotesize ETH Zurich\\
{\fontsize{8}{9}\selectfont \textsf{andrey@vision.ee.ethz.ch}}
\and
Radu Timofte\\
\footnotesize ETH Zurich\\
{\fontsize{8}{9}\selectfont \textsf{timofter@vision.ee.ethz.ch}}
\and
William Chou\\
\footnotesize Qualcomm, Inc.\\
{\fontsize{8}{9}\selectfont \textsf{wchou@qti.qualcomm.com}}
\and
Ke Wang\\
\footnotesize Huawei, Inc.\\
{\fontsize{8}{9}\selectfont \textsf{michael.wangke@huawei.com}}
\and
Max Wu\\
\footnotesize MediaTek, Inc.\\
{\fontsize{8}{9}\selectfont \textsf{max.wu@mediatek.com}}
\and
Tim Hartley\\
\footnotesize Arm, Inc.\\
{\fontsize{8}{9}\selectfont \textsf{tim.hartley@arm.com}\vspace{2mm}}
\and
Luc Van Gool \thanks{We also thank Przemyslaw Szczepaniak (pszczepaniak@google.com), Google Inc., for writing and editing sections 2.7, 3.1 and 3.2.}\\
\footnotesize ETH Zurich\\
{\fontsize{8}{9}\selectfont \textsf{vangool@vision.ee.ethz.ch}}
}

\date{}

\maketitle

\begin{abstract}
Over the last years, the computational power of mobile devices such as smartphones and tablets has grown dramatically, reaching the level of desktop computers available not long ago.
While standard smartphone apps are no longer a problem for them, there is still a group of tasks that can easily challenge even high-end devices, namely running artificial intelligence algorithms.
In this paper, we present a study of the current state of deep learning in the Android ecosystem and describe available frameworks, programming models and the limitations of running AI on smartphones. We give an overview of the hardware acceleration resources available on four main mobile chipset platforms: Qualcomm, HiSilicon, MediaTek and Samsung. Additionally, we present the real-world performance results of different mobile SoCs collected with AI Benchmark\footnote{\url{http://ai-benchmark.com}} that are covering all main existing hardware configurations.
\end{abstract}

\section{Introduction}

With the recent advances in mobile system-on-chip (SoC) technologies, the performance of portable Android devices has increased by a multiple over the past years. With their multi-core processors, dedicated GPUs, and gigabytes of RAM, the capabilities of current smartphones have already gone far beyond running the standard built-in phone applications or simple mobile games. Whereas their computational power already significantly exceeds the needs of most everyday use cases, artificial intelligence algorithms still remain challenging even for high-end smartphones and tablets. Despite the fact that many machine learning solutions are highly useful when deployed on end-user devices, running them on mobile platforms is associated with a huge computational overhead on phone CPUs and a serious drain on battery power.

Many recent developments in deep learning are, however, tightly connected to tasks meant for mobile devices. One notable group of such tasks is concerned with computer vision problems like image classification~\cite{krizhevsky2012imagenet,szegedy2016rethinking,howard2017mobilenets}, image enhancement~\cite{ignatov2017dslr,ignatov2017wespe,ignatov2018pirm} and super-resolution~\cite{dong2016image,ledig2017photo,timofte2018ntire}, optical character recognition~\cite{netzer2011reading}, object tracking~\cite{wu2015object,huang2017speed}, visual scene understanding~\cite{li2009towards,cordts2016cityscapes}, face detection and recognition~\cite{li2015convolutional,schroff2015facenet}, gaze tracking~\cite{zhang2015appearance}, etc. Another group of tasks encompasses various natural language processing problems such as natural language translation~\cite{sutskever2014sequence,bahdanau2014neural}, sentence completion~\cite{mikolov2013efficient,hu2014convolutional}, sentence sentiment analysis~\cite{socher2013recursive,severyn2015twitter} or interactive chatbots~\cite{serban2017deep}. A separte group deals with on-line sensor data processing for human activity recognition from accelerometer data~\cite{kwapisz2011activity,ignatov2018real}, gesture recognition~\cite{ordonez2016deep} or sleep monitoring~\cite{sathyanarayana2016sleep}. Several other deep learning problems on smartphones are related to speech recognition, virtual reality and many other tasks.

\begin{figure*}[t!]
\centering
\resizebox{0.8\linewidth}{!}
{
\includegraphics[width=1.0\linewidth]{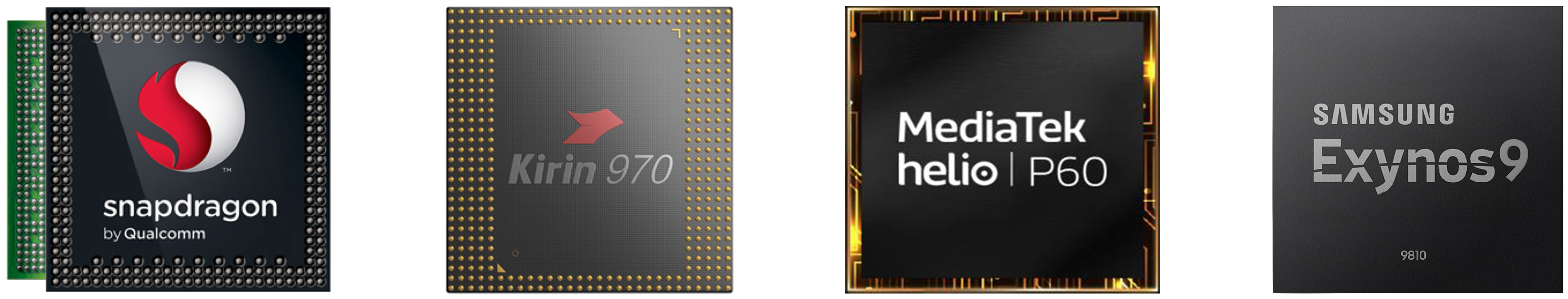}
}
\vspace{1.6mm}
\caption{\small{Mobile SoCs with potential acceleration support for third-party AI applications.}}
\vspace{-1.2mm}
\label{fig:chipsets}
\end{figure*}

Despite the rising interest in deep learning for mobile applications, the majority of AI algorithms are either not available on smartphones or are executed on remote servers due to the aforementioned phones' hardware limitations. The latter option is also not flawless, causing: a) privacy issues; b) dependency on an internet connection; c) delays associated with network latency; d) bottleneck problems~--- the number of possible clients depends on the servers' computational capabilities. To overcome these issues, there were a number of attempts to port separate algorithms or whole machine learning libraries to mobile platforms with added hardware acceleration (HA) using GPUs or DSPs. In~\cite{lane2015can}, the authors implemented a mobile neural network classification engine capable of sensor inference tasks on Qualcomm's Hexagon DSP~\cite{codrescu2014hexagon}. Though they achieved very impressive energy consumption results, the DSP was able to run only very simple CNN models due to its small program and memory space. In~\cite{latifi2016cnndroid}, the authors presented a GPU-accelerated library CNNdroid for parallel execution of pre-trained CNNs on mobile GPUs. The library was based on the RenderScript framework~\cite{guihot2012renderscript} that parallelizes computations across CPUs and GPUs, and though the proposed solution was up to 40 times faster compared to the baseline naive singe-thread implementation, in reality its speed was comparable to a CPU-based TensorFlow Mobile library~\cite{TensorFlowMobile2018} relying on the Arm NEON~\cite{reddy2008neon} instruction set. Motamedi \textit{et al.}~\cite{motamedi2018cappuccino} exploited the same approach of using RenderScript, but used a CPU's imprecise computing modes to lower execution times. Despite the promising results, the effect inexact arithmetic had on accuracy was not investigated in depth in this paper, and therefore the applicability of this approach remains unclear. RSTensorFlow~\cite{alzantot2017rstensorflow} is another attempt to expoit RenderScript for GPU-based acceleration of matrix operations, and in this case it was used to directly modify the TensorFlow Mobile library. The results demonstrated that, while matrix multiplications can be executed up to 3 times faster, it is not possible to speed up the convolutional operations that take approximately 75\% of the total inference time. Additionally, the experiment revealed that RenderScript is not always using GPUs on all the devices~--- sometimes it is running on a CPU only, leading to slower execution times even compared to the original TF implementation.

Besides that, some SDKs for running computationally intensive operations were proposed directly by SoC manufacturers. In 2016, Qualcomm introduced the Snapdragon Neural Processing Engine (SNPE)~\cite{SNPE2018} to accelerate the execution of neural networks with their GPUs and DSPs. The next year HiSilicon proposed the HiAI platform~\cite{HiAI2018} for running neural networks on Kirin's NPU, and later \mbox{MediaTek} presented the NeuroPilot SDK~\cite{lee2018techology} that can trigger GPUs or APUs to run deep learning models. The biggest issue is that all these SDKs were developed for the corresponding chipsets only, \ie the application relying on HiAI will not run on Qualcomm SoC, and vice versa, thus forcing developers to create several versions of their app for each platform, or to give up on some of them. This situation changed with the introduction of the Android Neural Networks API (NNAPI)~\cite{NNAPI2018}, designed to run deep learning models on mobile devices. This API is basically an intermediate layer between the higher-level machine learning framework and the device's hardware acceleration resources, and is responsible for their communication and for scheduling the execution of tasks on the most suitable hardware. NNAPI still requires specific SoC vendors' drivers in order to run the computations on anything but a CPU, and therefore its default presence in Android 8.1+ does not automatically guarantee hardware acceleration support.

While there exists a number of common benchmarks testing the CPU and GPU performance of mobile phones, none of them measure the speed and acceleration of AI operations that can be achieved due to available AI chips and DSPs. In this paper, we present an AI Benchmark designed specifically to test the machine learning performance, available hardware AI accelerators, chipset drivers, and memory limitations of the current Android devices. It consists of a number of computer vision AI tests that are executed directly on the phones' hardware and that cover relevant deep learning architectures and operations. We provide a detailed description of the actual chipset platforms and popular mobile machine learning frameworks, and describe the limitations of running deep learning algorithms on smartphones. Finally, we present the in-the-wild performance of about 200 Android devices and major mobile chipsets, as collected with our AI Benchmark, for over 10,000 smartphones and tablets.

The rest of the paper is arranged as follows. In Section~2 we describe the hardware acceleration resources available on the main chipset platforms, as well as the programming interfaces for accessing them. Section~3 gives an overview of popular mobile deep learning frameworks. Section~4 provides a detailed description of the benchmark architecture, its programming implementation, and the computer vision tests that it includes. Section~5 shows the experimental results and inference times for different deep learning architectures,  for various Android devices and chipsets. Section~6 analyzes the obtained results. Finally, Section~7 concludes the paper.

\section{Hardware Acceleration}

While the first consumer computers were mostly equipped with a single, stand-alone CPU, it soon became clear that its computational performance is too limited for a number of multimedia applications. This led to the creation of special co-processors working in parallel with the main CPU. Their architecture was optimized for many signal processing tasks. The era of digital signal processors (DSPs) began in the early 1980s with the introduction of the NEC µPD7720~\cite{chance1990devices}, the AT\&T DSP1~\cite{hesseldahl1999legacy} and the TI TMS32010~\cite{guttag1996tms320c8x} co-processors. They established general principles of the DSP architecture used until now~\cite{hays2004dsps}: Harvard architecture, hardware block for multiply-accumulate (MAC) operations, VLIW and SIMD instruction sets for parallel computations, etc. Though the first DSPs had quite restricted capabilities due to their limited set of instructions and memory constraints, they were widely used till the mid 90s of the last century. They were popular for applications related to computer graphics, sound and video decoding, as mathematical co-processors and accelerators for various photo editing software, and even for running the first deep learning OCR models designed in 1989~\cite{lecun1989backpropagation}. The latter task of classifying handwritten digits using CNNs reached high speeds at that time (12 images per second) due to the efficient vector and matrix-based calculations. These resulted from the highly parallelizable DSP architectures and the hardware implementation of MAC operations. At the end of the 90s the popularity of DSPs started to decrease and in the consumer PC sector they were largely replaced by general-purpose CPUs with integrated DSP instructions, GPUs for efficient parallel computations, and FPGAs configurable for various specific problems.

At the beginning of the 1990s, DSPs started to appear in mobile phones. At first, they were used only for voice coding and compression, as well as for some radio signal processing. Later on, with the integration of cameras and many multimedia features like music and video playback in mobile devices, the integrated DSPs started to be extensively used for image, video and sound processing. In contrast to what happened with desktop computers, DSPs were not displaced here by CPUs and GPUs because they often offered superior performance at lower power consumption, so critical for portable devices. In recent years, the computational power of mobile DSPs and other SoC components has grown drastically, and now, complemented by GPUs, NPUs and dedicated AI cores, they enable AI and deep learning-based computations. A detailed description of the current mobile platforms (fig.~\ref{fig:chipsets}) and their hardware acceleration resources is provided below.

\subsection{Qualcomm chipsets / SNPE SDK}

\begin{figure}[t!]
\centering
\resizebox{0.9\linewidth}{!}
{
\includegraphics[width=1.0\linewidth]{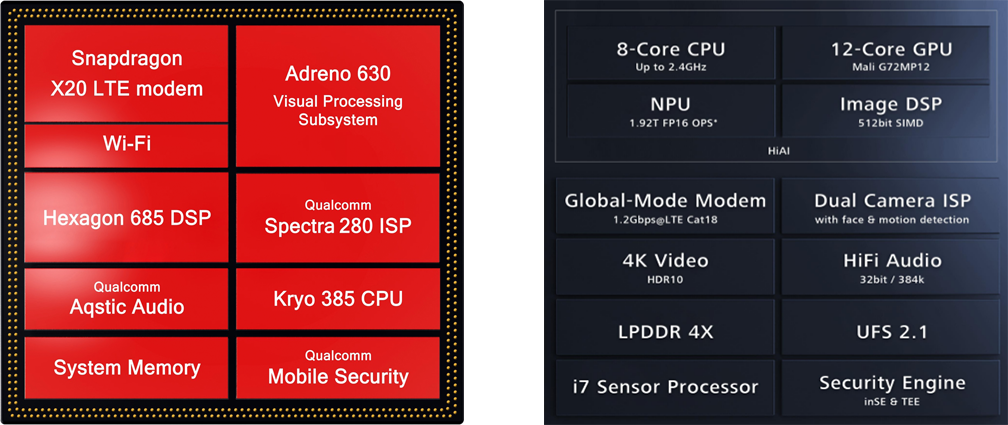}
}
\caption{\small{SoC components integrated into Snapdragon 845 (left) and Kirin 970 (right) chipsets.}}
\label{fig:soc-components}
\end{figure}

Qualcomm is an American semiconductor and wireless telecommunications company, founded in 1985. Its first Snapdragon mobile SoC QSD8250 was released in 2007 and already featured a dedicated AMD Z430 GPU and the first commercial generation of QDSP6 Hexagon DSPs. In 2009, after the acquisition of AMD's mobile graphics division, the corresponding GPU series was renamed to Adreno (anagram from Radeon), and its successors are present under this name in all current Snapdragon SoCs. Their performance evolved from 2.1 (Adreno 200) to 727 (Adreno 630) GFLOPS. The DSP architecture has also undergone significant changes from the first (2006) to the current sixth generation, and is now supporting wide vector extensions (HVX), dynamic multi-threading, VLIW and SIMD instruction sets. They can also be programmed by users~\cite{codrescu2014hexagon}. The main Snapdragon CPU cores have an Arm-based architecture and usually feature Qualcomm's own customized in-house design, often developed based on Arm Cortex cores. These three components (CPUs with the Arm NEON instruction set, GPUs and DSPs) form Snapdragon's heterogeneous computing architecture (fig.~\ref{fig:soc-components}) well suitable for running various AI algorithms. The Qualcomm chipsets are now covering around 55\% of the smartphone SoC market and are installed in many popular smartphones, tablets, and wearables.

Qualcomm first addressed the problem of on-device AI inference hardware acceleration in the Snapdragon 820 in May 2015 and also announced its proprietary Snapdragon Neural Processing Engine (SNPE) SDK in May 2016, which offers runtime acceleration across all Snapdragon's processing components. The SDK supports common deep learning model frameworks, such as Caffe/Caffe2, TensorFlow, PyTorch, Chainer, MxNet, CNTK and PaddlePaddle via ONNX. It is designed to enable developers to run their own custom neural network models on various Qualcomm-powered devices. The SDK is supported on 17 Snapdragon mobile processors starting from premium (Snapdragon 845, 835, 820), high tier (Snapdragon 710, 670, 660, 652, 650, 653, 636, 632, 630, 626 and 625) as well as the mid-tier (Snapdragon 450, 439, 429). It also supports the Qualcomm Vision Intelligence Platform (QCS603 and QCS605), designed for efficient machine learning on IoT devices.

Qualcomm's first NNAPI driver for running quantized neural networks on Hexagon DSPs was introduced in the Android O-MR1, though it was not used in any commercial devices at that time and first appeared only later in the OnePlus 6 and Xiaomi Mi8 with the next Android version. In Android P, these drivers got additional support for running float models on the Adreno GPU. Yet, they are currently not present in the market. The considered NNAPI drivers are generally adopting hardware acceleration principles and implementation used in SNPE SDK. The differences mainly come from the restrictions of the current Android NNAPI specifications. Qualcomm delivers these drivers in the software images provided to its OEM customers, which then in turn determine when and how to include them to end devices: with their initial release or later over the air in subsequent software updates. As a result, their presence and actual version might vary significantly across the phones on the market.

\subsection{HiSilicon chipsets / Huawei HiAI SDK}

\begin{figure*}[t!]
\centering
\resizebox{1.0\linewidth}{!}
{
\includegraphics[width=1.0\linewidth]{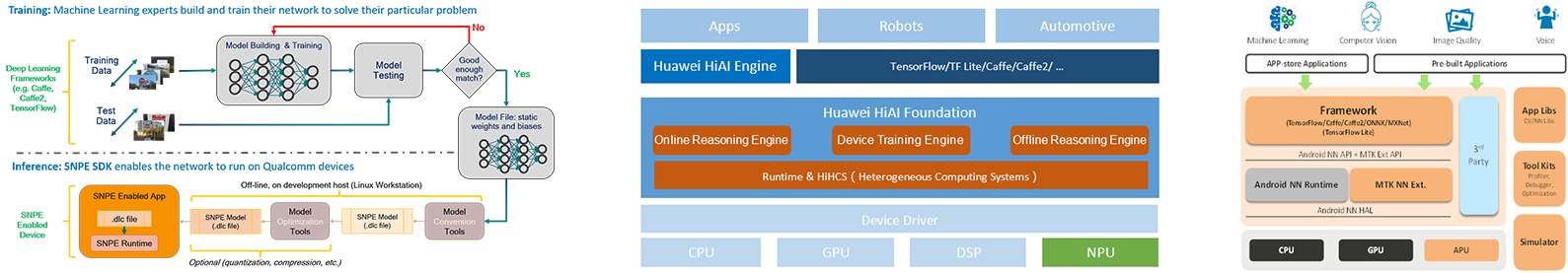}
}
\caption{\small{Schematic representation of SNPE, HiAI and NeuroPilot SDKs from Qualcomm, Huawei and MediaTek, respectively.}}
\label{fig:sdks}
\end{figure*}

HiSilicon is a Chinese semiconductor company founded in 2004 as a subsidiary of Huawei. Its first mobile processor (K3V1) was introduced in 2008, but the first commercially successful product used in a number of Android devices was the next SoC generation (K3V2) released in 2012 and featuring four Arm \mbox{Cortex-A9} CPU cores and a Vivante GPU. In 2014, a new Kirin SoC family consisting of mid-range (600 Series) and high-end (900 Series) chipsets was launched as a \mbox{successor} to the K3 series and is used in Huawei devices until now. Unlike Qualcomm, HiSilicon does not create customized CPU and GPU designs and all Kirin chipsets are based on off-the-shelf Arm Cortex CPU cores and various versions of Mali GPUs. A different approach was also developed for accelerating AI computations: instead of relying on GPUs and DSPs, HiSilicon introduced a specialized neural processing unit (NPU) aimed at fast vector and matrix-based computations widely used in AI and deep learning algorithms. According to Huawei, it delivers up to 25 times better performance and 50 times greater efficiency compared to the standard quad-core Cortex-A73 CPU cluster. The NPU design was licensed from the Cambricon Technologies company (\mbox{Cambricon-1A} chip) and is said to deliver a peak performance of about 1.92 TFLOPs, though this number mainly refers to quantized 8-bit computations. This NPU first appeared in the Kirin 970 SoC, and later two enhanced NPUs were also integrated into the subsequent Kirin 980 chipset. It should be noted that other SoCs apart from Kirin 970/980 do not contain this NPU module and are currently unable to provide acceleration for third-party AI-based applications. The aforementioned chipsets can be found only inside Huawei devices as they are not sold to external OEM companies; the current total market share of HiSilicon SoCs is around 10\%.

To give external access to Kirin's NPU, Huawei released in late 2017 the HiAI~\cite{HiAI2018} Mobile Computing Platform SDK, providing APIs for executing deep learning models on hardware resources integrated within Kirin SoC. This SDK is now supporting only Caffe, Tensorflow Mobile and Lite frameworks, though in future releases it might also offer support for Caffe2 and ONNX. It provides acceleration for 16-bit float, 8-bit and 1-bit quantized models, and can additionally speed-up sparse models by skipping multiply-add operations containing zero variables. Apart from low-level APIs, the HiAI Engine also provides a ready-to-use implementation of several computer vision algorithms including image categorization, face and facial attribute detection, document detection and correction, image super-resolution, QR code detection, etc.

Starting from Android 8.1 (EMUI 8.1), Huawei is including NNAPI drivers for its Kirin 970/980 chipsets that are generally based on the HiAI implementation. Currently, they are providing support only for 16-bit float models, quantized networks will be supported in the future releases. It should be mentioned that all Huawei devices that are based on other chipsets do not contain NNAPI drivers as they are lacking the above-mentioned NPU module.

\subsection{MediaTek chipsets / NeuroPilot SDK}

MediaTek is a Taiwanese semiconductor company spun off from the United Microelectronics Corporation in 1997. Its mobile division was launched in 2004 and soon after this MediaTek released its first mobile chipsets that were used in many entry-level Chinese phones and smartphones produced at that time. It gained popularity on the global smartphone market in 2013 with the introduction of the MediaTek 657x/658x family of dual and quad-core SoCs with Mali or PowerVR graphics, and later with the release of 64-bit MediaTek MT67xx chipsets they became widely used in many Android devices from various OEMs, getting a market share of about 20\%. Similarly to Huawei, MediaTek is integrating into its SoCs standard Arm Cortex CPU cores and Mali or \mbox{PowerVR} GPUs. At the beginning of 2018, MediaTek addressed the problem of accelerating machine learning-based applications by launching their Helio P60 platform with embedded AI processing unit (APU). This APU can deliver the performance of up to 280GMAC/s for 8-bit computations and is primarily used for accelerating quantized neural networks, while float models are running on four Cortex-A53 CPU cores and Mali-G72 MP3 GPU clocked at 800MHz. Thus, MediaTek's approach lies in between the solutions from Huawei and Qualcomm: a dedicated chip for quantized computations (as in Kirin's SoC) and CPU/GPU for float ones (as in Snapdragon chipsets).

The release of the Helio P60 was accompanied by the introduction of \mbox{MediaTek's} NeuroPilot SDK~\cite{lee2018techology} constructed around TensorFlow Lite and Android NNAPI. This SDK consists of four main components: 1) TOCO-based tools for quantizing float TF Lite networks and for converting pre-trained TensorFlow/Caffe/ONNX models (with supported operations) to TensorFlow Lite format. 2) An extended list of implemented TF Lite operations and the corresponding interpreter for loading and running converted .tflite models. 3) APU and GPU NNAPI drivers implementing hardware accelerated operations for MediaTek's NeuroPilot platform; the APU drivers currently only support INT8 ops and GPU drivers~--- FP16/32 ops. 4) Facilities for profiling and debugging neural network-based applications, and an interface for pinning target operations on a specific hardware accelerator like GPU or APU. The SDK is supporting purely MediaTek NeuroPilot-compatible chipsets (currently Helio P60 only).

There also exists a corresponding stand-alone version of NNAPI drivers supporting float and quantized models. Nonetheless, except for the P60 developer platform, only one commercial device with MediaTek P60 chipset (Vivo V11) is known to contain these drivers.

\subsection{Samsung chipsets}

Samsung Electronics is a South Korean electronics company founded in 1969. In 1988, it merged with Samsung Semiconductor \& Communications and obtained its current name. That same year it launched its first mobile phone, while its first mobile processor (S3C44B0, 66 MHz, Armv4) was presented only in 2000. Later it significantly extended its S3Cxxxx and S5Pxxxx SoC series that were widely used in many Windows Mobile devices, in the iPhone 2G/3/3GS, and in some early Android smartphones. With the introduction of the S5PC110 chipset in 2010, all Samsung SoCs were rebranded into Exynos and are using this name up to now (Exynos 3-9th generations). Similarly to Huawei and MediaTek, Samsung is primarily using Arm Cortex CPU cores and Mali or PowerVR graphics in its chipsets, though starting from Exynos 8 it is also integrating its in-house developed Mongoose Arm-based CPU cores into high-end SoCs. As for specific AI chips, Samsung introduced in the Exynos 8895 a Vision Processing Unit (VPU) mainly used by its phones' cameras. Yet, no drivers, SDKs or additional details were released, making it inaccessible by third-party applications. Only two Samsung devices (Note 9 and Tab S4) are currently running Android 8.1+ and are using Google's default NNAPI drivers utilizing the CPU only. According to some rumors, the next Exynos chipset might include a dedicated AI chip, though this information was not officially confirmed by Samsung. The current market share of Samsung chipsets is around 10\%.

\subsection{Google Pixel / Pixel Visual Core}

\begin{figure}[t!]
\centering
\resizebox{0.6\linewidth}{!}
{
\includegraphics[width=1.0\linewidth]{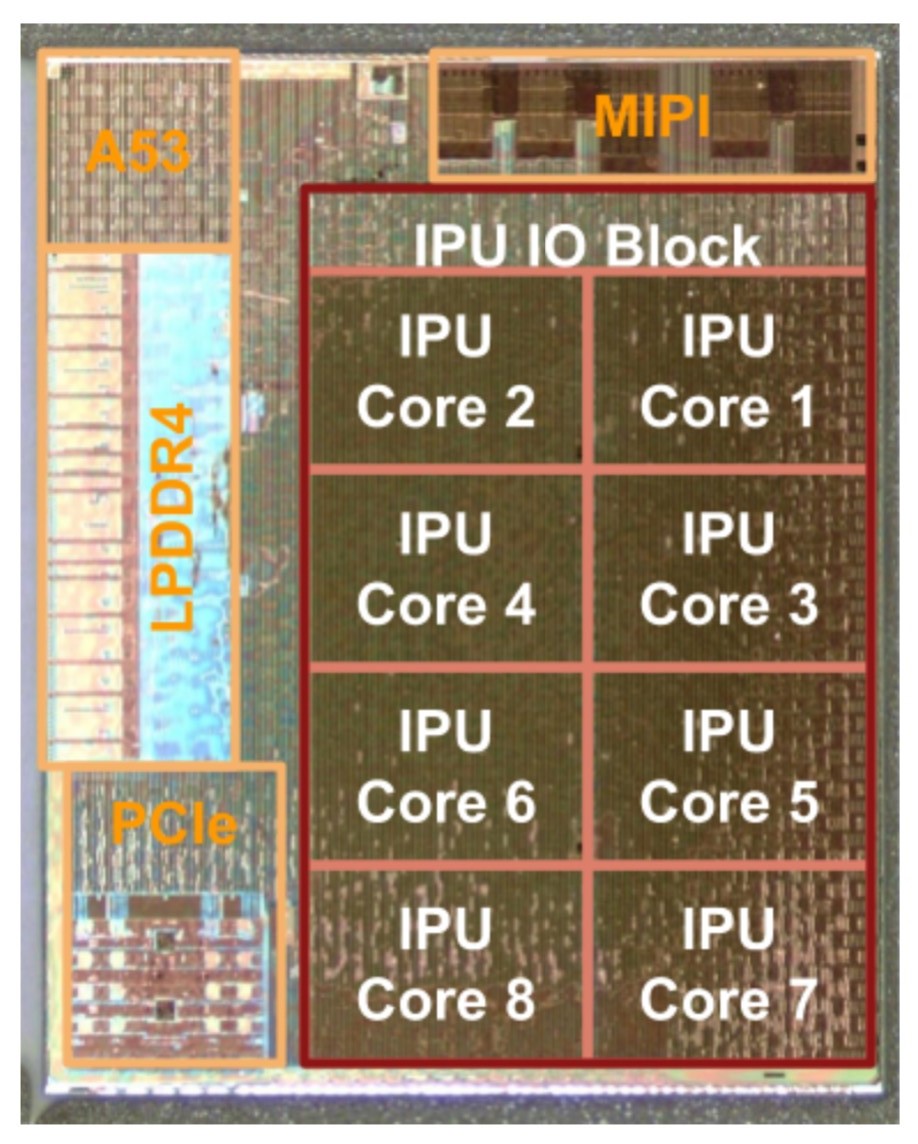}
}
\caption{\small{The architecture of the Pixel Visual Core AI Chip.}}
\label{fig:pixel-core}
\end{figure}

Apart from its Android operating system, Google started, since Android 2.1, to annually release smartphones and tablets under the Google Nexus brand. These were developed in collaboration with external OEMs, among which at different times were HTC, Samsung, LG, Motorola, Huawei and Asus. These devices were featuring the stock Android operating system running on the latest high-end hardware and were the first to receive Android updates (with the possibility of installing beta versions). In 2016 the Nexus product line was discontinued and all new smartphones started being produced under the Google Pixel brand, though the aforementioned principles remained the same. The majority of these devices were based on Qualcomm chipsets, therefore all information from the above Qualcomm section can be applied to them too. Yet, starting from Pixel 2 (XL), Google has added to its smartphones a dedicated fully-programmable Pixel Visual Core AI chip (fig.~\ref{fig:pixel-core}), separate from the main Qualcomm SoC and developed in collaboration with Intel. The chip contains one Arm Cortex-A53 core for handling communications with the main application processor, integrated LPDDR4 RAM and eight custom image processing unit (IPU) cores. Each IPU contains 512 arithmetic logic units with 256 processing elements arranged as a 16$\times$16 two-dimensional array and supports a custom VLIW instruction set. The chip provides native support for 8-bit and 16-bit integer computations and delivers a performance of up to 3.2 TFLOPS. Although the Pixel Visual Core is generally compliant with TensorFlow (Lite), Google did not release the corresponding SDK and NNAPI drivers, thus it cannot be used by external developers for accelerating machine learning-based applications and its present use is mainly limited to Google's HDR+ image processing.

\subsection{Arm Cortex CPUs / Mali GPUs / NN SDK}

Currently, all CPU cores integrated into mobile SoCs are based on the Arm architecture, and in devices not supporting HA for machine learning applications these CPUs are responsible for running all AI algorithms. To speed-up the computations in this case, Arm has introduced a number of specific instruction sets aimed at fast vector- and matrix-based calculations. The most notable technology here is the Arm NEON\cite{reddy2008neon}~--- an advanced SIMD (single instruction multiple data) architecture extension first introduced in Armv7 processors. NEON basically implements DSP-like instructions for concurrent computations and allows the simultaneous execution of up to 16x8-bit, 8x16-bit, 4x32-bit, 2x64-bit integer and 8x16-bit, 4x32-bit, 2x64-bit floating-point operations. Additionally, Arm has recently presented its new DynamIQ technology that is able to efficiently utilize all cores within a single Arm CPU for parallel computations, and a specific instruction for calculating dot products in the Armv8.4-A microarchitecture. Many of these optimized instructions are integrated in Google's default NNAPI drivers, handling the CPU path when no other means for acceleration are available.

Apart from that, Arm has also presented the Arm NN SDK~\cite{ArmNN2018} to accelerate machine learning computations on mobile SoCs. It provides both the CPU and GPU paths for ML workloads, along with parsers for TensorFlow, Caffe, ONNX and TFLite. On the CPU side it is compatible with any platform with Armv7 and above CPUs (assuming NEON availability), with key low level optimizations for specific architectures. The GPU path will be available on platforms with Arm Mali GPUs, either from the Midgard family (Mali-T6xx and onwards when GPGPU was introduced) or the later Bifrost family (G71 / G51 and onwards), and requires the Mali GPU and OpenCL drivers to be installed. The Arm NN SDK provides support for both FP32 and quantized INT8 networks and can run on Linux or Android platforms in parallel to NNAPI.

\subsection{Android NNAPI}

\begin{figure}[t!]
\centering
\resizebox{0.8\linewidth}{!}
{
\includegraphics[width=1.0\linewidth]{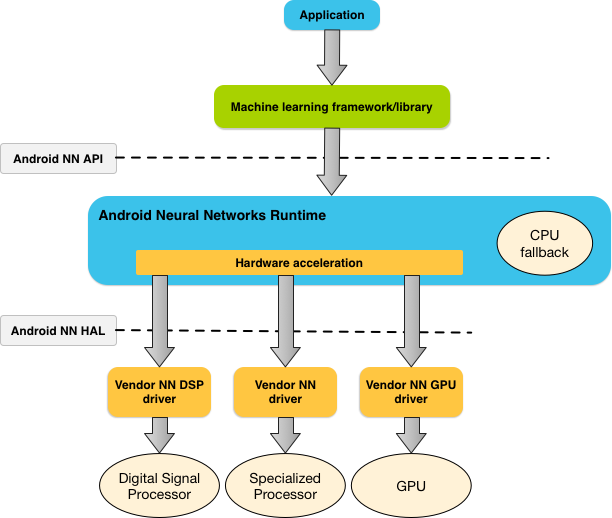}
}
\caption{\small{System architecture for Android Neural Networks API.}}
\label{fig:nnapi}
\end{figure}

While there exist a number of proprietary SDKs for accessing DSPs, GPUs or NPUs on different mobile platforms, this was not really solving the problem of using HA for running deep learning algorithms on mobiles, as all these SDKs are providing access only to some particular chipsets and are additionally incompatible with each other. To solve this problem, Google has recently introduced a unified Android Neural Networks API (NNAPI) that is an Android C API designed for running computationally intensive machine and deep learning operations on mobile devices. The system architecture of NNAPI is presented in the figure~\ref{fig:nnapi}. Apps typically would not use NNAPI directly, instead they will rely on higher-level machine learning frameworks that in turn could use NNAPI to run hardware-accelerated inference on supported devices. To perform computations using NNAPI, the executed model should be first represented as a directed graph that defines the computations to perform. This graph, combined with the data defining the model (\eg the weights and biases passed down from a machine learning framework), forms the model for NNAPI runtime evaluation. Based on the app's requirements and device hardware, Android's neural networks runtime can efficiently distribute the computation workload across available on-device processors, including dedicated neural network chips, GPUs and DSPs. NNAPI is available on all devices running Android 8.1 (API level 27) or higher, but it still requires a specialized vendor driver for accessing the device's hardware. For devices that lack this driver, the NNAPI runtime relies on optimized code to execute requests on the CPU.

\section{Deep Learning Mobile Frameworks}

With the widespread use of the Android operating system, a number of popular deep learning frameworks were ported to this platform, including Torch~\cite{TorchAndroid2018}, Deeplearning4j~\cite{Deeplearning4j2018}, TensorFlow (Mobile~\cite{TensorFlowMobile2018}, Lite~\cite{TensorFlowLite2018}), Caffe~\cite{CaffeAndroid2018}, Caffe2~\cite{Caffe2Android2018}, MXNet~\cite{MXNet2018}, NNabla~\cite{NNabla2018}, etc. Nowadays, the most commonly used are three of them: Tensorflow Mobile, Tensorflow Lite and Caffe2 that are described below.

\begin{figure*}[t!]
\centering
\setlength{\tabcolsep}{1pt}

\begin{subfigure}[t]{.49\linewidth}
\centering\includegraphics[width=1.0\linewidth]{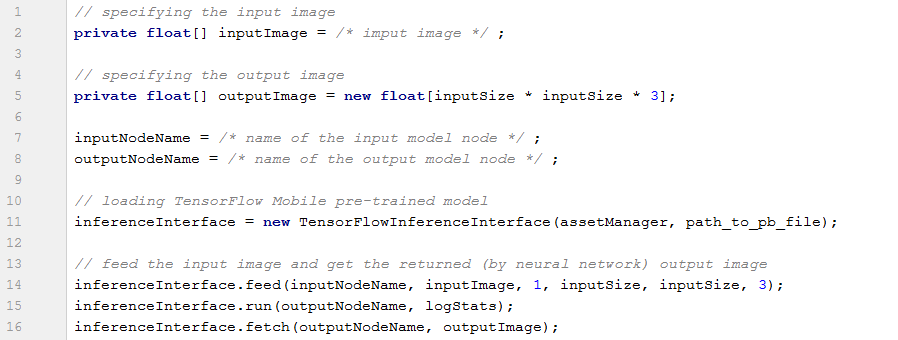}
\caption{\footnotesize{TensorFlow Mobile}}
\end{subfigure}
\begin{subfigure}[t]{.49\linewidth}
\centering\includegraphics[width=1.0\linewidth]{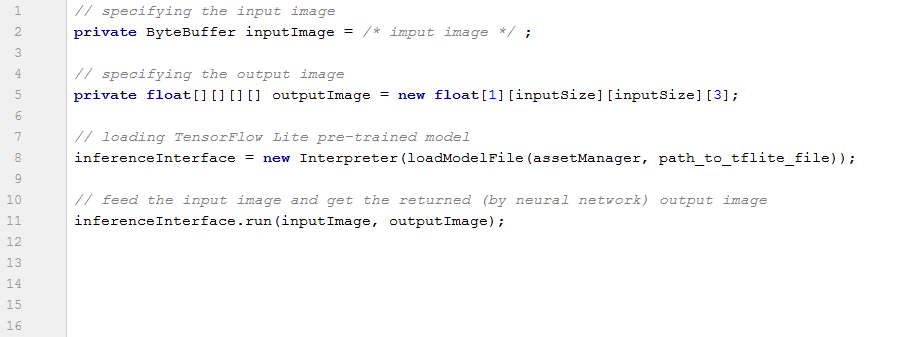}
\caption{\footnotesize{TensorFlow Lite}}
\end{subfigure}
\caption{\small{Code snippets of TensorFlow Mobile and Lite Android Java interfaces.}}
\vspace{-0.7mm}
\label{fig:tfcode}
\end{figure*}

\subsection{TensorFlow Mobile}

Tensorflow~\cite{abadi2016tensorflow} is an open-source machine learning library for research and development released by Google in 2015. TensorFlow's programming model can be described as a directed graph that defines the relation between the input and output (target) variables. The graph itself consists of a set of nodes representing various operators applied sequentially to the input data (\eg convolutional, pooling, LSTM layers, etc.) that are defining a deep learning model and the corresponding dataflow computation. After the model is trained, it can be exported as a .pb graph and executed on mobile devices using the TensorFlow Mobile library~\cite{TensorFlowMobile2018}, available on Android as well as iOS platforms. A code snippet of the corresponding Java inference interface is presented in figure~\ref{fig:tfcode} (a). Note that there is no need to specify the model architecture in the actual application code: it is already stored along with pre-trained weights in the .pb graph, and developers only need to provide the location of this file and the input data.

The main advantage of the TensorFlow Mobile library is that it supports the majority of operations available in the standard TF version, therefore almost any TensorFlow model can be converted and executed on a mobile device. Additionally, all current SDKs from SoC manufacturers (SNPE~\cite{SNPE2018}, HiAI~\cite{HiAI2018}, NeuroPilot~\cite{lee2018techology} and ArmNN~\cite{ArmNN2018}) are providing (partial) hardware acceleration support for this library. This said, the development of TensorFlow Mobile is coming to a close, as Google announced its gradual deprecation in favor of the TensorFlow Lite library~\cite{TensorFlowMobileVsLite2018}. Particularly, TF Mobile will not get Android NNAPI support, thus without using specific SDKs all models will still be executed on CPUs only.

\subsection{TensorFlow Lite}

TensorFlow Lite~\cite{TensorFlowLite2018} was presented late 2017, as a successor of the TF Mobile library. According to Google, it provides better performance and a smaller binary size due to optimized kernels, pre-fused activations and fewer dependencies. Similarly to TF Mobile, a general TensorFlow pre-trained model can be in theory converted to .tflite format and later used for inference on Android or iOS platforms, the corresponding Java code snippet is shown in figure~\ref{fig:tfcode} (b). The change of the file format (.tflite instead of .pb) is caused by the use of a new FlatBuffers serialization library that allows to access saved models without a parsing/unpacking step, often coupled with per-object memory allocation. Finally, the new library is compatible with Android NNAPI and can by default run with hardware acceleration on devices with appropriate chipsets and drivers.

It should be noted, however, that TensorFlow Lite is in developer preview at the moment and has a number of substantial limitations. First of all, it supports only a limited set of operators, lacking the full support of, \eg image resizing, batch and instance normalization, LSTM units, some statistical functions or even simple mathematical operations like exponentiation or argmax. Officially, Google guarantees only three models to work: the Inception-V3, MobileNet and Smart Reply SSL algorithm, though with some modifications it is possible to run a number of other deep learning models. A second issue concerns the inference time and the amount of consumed RAM. Since the ByteBuffer format is not supported for the network's output, these two values can be up to 2$\times$ higher compared to TF Mobile for image-to-image translation problems. Finally, stability is another concern~--- the current official version might not work flawlessly with a number of models and mobile devices, though some of the issues are already solved in the nightly TF Lite version. While many of these problems will probably be overcome in the upcoming library releases, currently they make the use of TensorFlow Lite complicated for many existing deep learning problems.

\subsection{Caffe2}

Caffe~\cite{jia2014caffe} is another open-source deep learning framework, originally developed at UC Berkeley by Yangqing Jia and released in 2013. Its first unofficial Android port appeared the next year~\cite{CaffeAndroid2018}, and in 2017, with Facebook's release of the successor, Caffe2, its mobile version for iOS and Android platforms was also presented~\cite{Caffe2Android2018}. Caffe2 is using a programming model similar to TensorFlow's, with static computational graphs and nodes representing various operators.
According to the Caffe2 github repository~\cite{Caffe2AICamera2018}, the speed of its mobile library is generally comparable to that of TensorFlow Lite~\cite{TFLiteGoogleBenchmark2018} (175ms vs. 158ms for the SqueezeNet model on Snapdragon 821 SoC). Report~\cite{Caffe2onAndroidPresentation2018} additionally claims about up to a 6x speed-up when using the OpenGL backend for GPU-based computations, but this feature is not yet available in the current Caffe2 release. Similarly to TensorFlow, acceleration for Caffe2 models is also supported by all proprietary SDKs (SNPE, HiAI, NeuroPilot and ArmNN), but NNAPI support is still in development and is not fully integrated yet.

\section{AI Benchmark}

The AI Benchmark is an Android application designed to check the performance and the memory limitations associated with running AI and deep learning algorithms on mobile platforms. It consists of several computer vision tasks performed by neural networks that are running directly on Android devices. The considered networks represent the most popular and commonly used architectures that can be currently deployed on smartphones, their detailed description along with technical details of the application are provided below.

\subsection{Deep Learning Tests}

The actual benchmark version [2.0.0] consists of the following nine deep learning tests.

\smallskip
\smallskip

\begin{figure*}[t!]
\centering
\resizebox{1.0\linewidth}{!}
{
\includegraphics[width=1.0\linewidth]{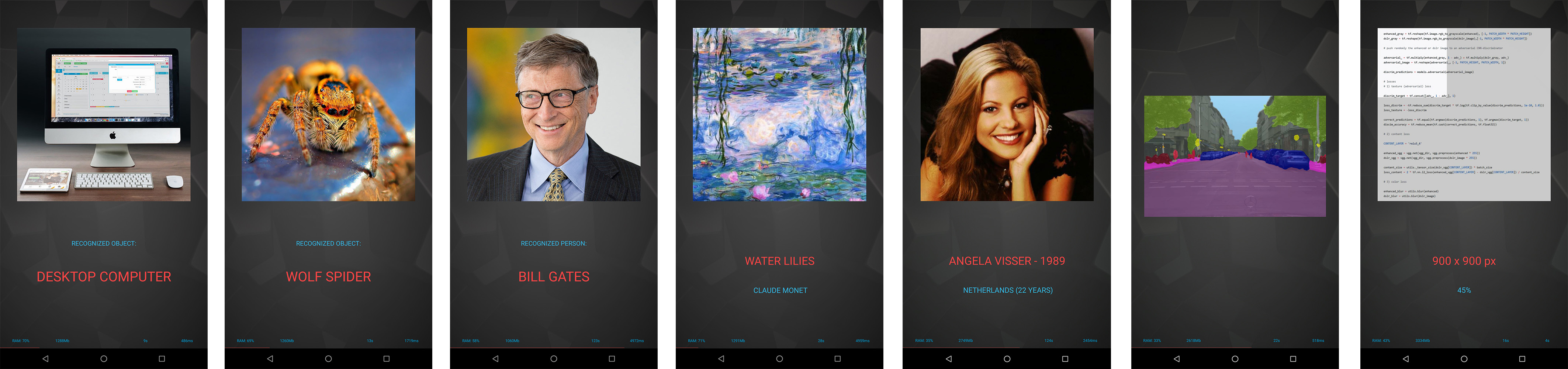}
}
\caption{\small{Sample result visualizations displayed to the user in the considered deep learning tests.}}
\label{fig:tests}
\end{figure*}

\noindent
\textbf{Test 1: Image Recognition.} This task represents a conventional ImageNet challenge where the goal is to classify images into 1000 categories. In the first test, classification is done with a resource-efficient MobileNet-V1~\cite{howard2017mobilenets} architecture designed specifically for mobile and embedded vision applications. The network mainly consists of 1$\times$1 convolutional (75\%) and fully connected (24\%) layers, where 95\% of the total 569M multiply-add operations happens in the first ones. MobileNet achieves 70.6\% accuracy on the ImageNet dataset, thus outperforming the larger AlexNet, SqueezeNet and Inception-V1 models. It can be optimized further for mobile usage by quantization~\cite{jacob2017quantization,sheng2018quantization}~--- converting its weights and activations from FLOAT32 to INT8 8-bit fixed point representation. Though this leads to an accuracy drop to 69.7\%, the speed is simultaneously more than doubled and the size is reduced (by a factor of 4) to 4.3MB. The latter quantized MobileNet-V1 is deployed in the first test.

\smallskip
\smallskip

\noindent
\textbf{Test 2: Image Recognition.} The same ImageNet classification problem as above, but in the second test a considerably larger and more accurate Inception-V3~\cite{szegedy2016rethinking} CNN, presented by Google in 2015, is used. This network is comprised of 11 inception blocks that mainly consist of 1$\times$1, 1$\times$3  + 3$\times$1, 1$\times$7 + 7$\times$1 and 3$\times$3 convolutional layers. In contrast to MobileNet, Inception-V3 requires about 5,000M multiply-add operations, and the size of the saved CNN is around 96MB. The accuracy is significantly higher too however,~--- 78\% on the same ImageNet dataset and currently the best result among popular networks of size below 100MB.

\smallskip
\smallskip

\noindent
\textbf{Test 3: Face Recognition.} The goal of this task is to retrieve the most similar face to a given one from an existing facial database. To do this, a neural network is first trained to produce a small feature vector for each facial image that encodes its visual features and is invariant to face scaling, shifts and rotations. In this test, we are using the Inception-Resnet-V1 network~\cite{szegedy2017inception}, presented by Google in 2017. It was trained to minimize a triplet loss~\cite{schroff2015facenet} on the VGGFace2 dataset~\cite{FaceNetgithub2018}.
After the network is trained, it is applied to a new facial image and produces its feature vector that is then used to retrieve the closest vector (and the respective identity) from the database. The size of the input images in this task is 512$\times$512 pixels, and the dimensionality of feature vectors is 128. The architecture of the Inception-ResNet-V1 consists of 20 inception blocks and is conceptually similar to the previously discussed Inception-V3 CNN; their size, accuracy on the ImageNet dataset, and computational cost are very similar as well. The biggest benefit of this network is its training speed~--- it needs fewer epochs to achieve the same accuracy than Inception-V3.

\smallskip
\smallskip

We would like to note that the models used in the first three tests currently represent a core set of architectures for classification problems that are suitable for mobile deployment. Networks faster than MobileNet (or its variants) are showing substantially worse accuracy. Models with better precision than Inception-V3 or Inception-ResNet-V1 have sizes exceeding 100-150MB~\cite{TFSlim2018}, which makes their application on mobile devices quite complicated due to the resulting size of the APK file. Quantization of these networks can partially solve the problem, but currently their quantized versions are not yet publicly available.

\smallskip
\smallskip

\noindent
\textbf{Test 4: Image Deblurring.} This test is aimed at removing Gaussian blur from images, which is done using the SRCNN network~\cite{dong2016image}~--- one of the first CNNs proposed for the super-resolution problem that is now widely used as a baseline for many image-to-image translation tasks. The architecture of this network is very shallow: three layers with 9$\times$9 and 5$\times$5 filters, in total 69,162 parameters and around 64B multiply-add operations for HD-resolution image. As a result, the size of the saved pre-trained network is only 278KB.

\smallskip
\smallskip

\noindent
\textbf{Test 5: Image Super-Resolution.} The goal of the super-resolution task is to reconstruct the original image from its downscaled version. In this test we consider a downscaling factor of 3, and image restoration is performed by the VDSR~\cite{kim2016accurate} network, presented in 2015 shortly after SRCNN. This network features a VGG-based architecture that is composed of 19 convolutional layers with 3$\times$3 filters, enough to obtain top quantitative results on many image processing problems. The VDSR network has 665K parameters and requires around 600B multiply-add operations for HD images; the size of the network is 2.7MB.

\smallskip
\smallskip

\noindent
\textbf{Test 6: Image Super-Resolution.} This test solves the same super-resolution problem, but with a downscaling factor of 4 and using the SRGAN~\cite{ledig2017photo} model that consists of two neural networks. The first one is ResNet previously proposed in~\cite{johnson2016perceptual} that in this implementation consists of 16 residual blocks; this network performs image restoration. The second one is an adversarial CNN~--- it is trained to distinguish between the real high-resolution images and the images reconstructed by ResNet. During the training, these networks are playing the following game: the adversarial CNN is trying to maximize its classification accuracy, while ResNet has the opposite goal of minimizing it, \ie to provide reconstructed images that are indistinguishable from the target ones. In practice, this leads to much better perceptual results than when using the standard Euclidean norm or content-based losses. After the model is trained, the adversarial CNN is removed and inference is performed by ResNet only. The latter network contains 1.5M parameters and the size of the saved pre-trained model is 6.2MB.

\smallskip
\smallskip

\noindent
\textbf{Test 7: Image Semantic Segmentation.} In contrast to image classification, the goal of this task is to get a pixel-level image understanding, meaning that each pixel has to be classified as belonging to one of 19 categories: car, pedestrian, road, sky, vegetation, etc. This is done with an ICNet CNN~\cite{zhao2017icnet}, designed for fast and accurate segmentation on low-performance devices. The speedup was mainly achieved by downsampling and shrinking feature maps, though the resulting accuracy on the Cityscapes dataset remained high~--- 70.6\% mIoU. ICNet consists of 6.7M parameters and the size of the pre-trained model is 27MB.

\smallskip
\smallskip

\noindent
\textbf{Test 8: Image Enhancement.} We consider here a general image and photo enhancement problem that encompasses various kinds of improvements including color enhancement, denoising, sharpening, texture synthesis, etc. In this formulation the problem was first addressed in the DPED paper~\cite{ignatov2017dslr}, where the authors were trying to turn low-quality smartphone photos into photos as they would be taken with a DSLR camera. This work adopted a ResNet-like architecture with 4 residual blocks and proposed specific losses targeted at various aspects of image quality. The obtained results demonstrated superior visual quality compared to the results of manual retouching or standard automatic algorithms. The main limitation of the approach was a need of device-specific training. The network is parameterized by 400K parameters and has a size of 1.6MB.

\smallskip
\smallskip

\noindent
\textbf{Test 9: Memory Limitations.} While previous tests were mainly evaluating the runtime of various deep learning models, the goal of the last test is to check RAM resources that can be allocated for running neural networks. In this test we are using the same SRCNN model as in the fourth task (deblurring), while gradually increasing the size of the input image until we run into a memory exception, meaning that the device does not have enough RAM to process larger inputs. The SRCNN model was chosen here since it consumes an amount of RAM similar to other models (for images of the same resolution), while its runtime is much faster and thus the test requires less time to finish. It is useful to note that the memory consumed by a network is primarily determined by the dimensions of its largest (convolutional) layer, which in the case of SRCNN is the first layer with 64 convolutional filters.

\smallskip
\smallskip

These nine tests represent the current deep learning core of the benchmark (fig.~\ref{fig:tests}); its technical components and implementation details are discussed below.

\subsection{Technical Description}

The current release of the AI Benchmark (2.0.0) is using the TensorFlow Lite~\cite{TensorFlowLite2018} library as a backend for running all embedded deep learning models.
Though the previous release was originally developed based on TF Mobile~\cite{TensorFlowMobile2018}, its lack of NNAPI support imposed critical constraints on using hardware acceleration resources, and thus was later deprecated.
The actual benchmark version was compiled with the latest TF Lite nightly build where some issues present in the stable TensorFlow versions were already solved.

The benchmark consists of nine deep learning tests described in the previous section. These can be generally divided into two groups. The first group includes tests 1, 2, 4, 5, 8, 9. Those use CNN models fully supported by NNAPI (\ie all underlying TensorFlow operations are implemented in NNAPI introduced in Android 8.1), and therefore they can run with hardware acceleration on devices with appropriate chipsets and drivers. NNAPI is always enabled in these tests to avoid the situation when the system fails to automatically detect the presence of AI accelerators and performs all computations on CPU. It should also be mentioned that the first test runs a quantized CNN model and is used to check the performance of accelerated INT8-based computations.

The second group contains the other three tests, \textit{i.e.} 3, 6 and 7, where neural networks are always running entirely on CPU. They contain at least one TF operation that is not yet present in NNAPI, and using partial acceleration for supported ops only is currently not possible. These tests were added to evaluate the speed of CPU-based execution and the performance of the Arm NEON instruction set~\cite{reddy2008neon}, present in all current Arm processors and designed specifically for high-performance computing and image processing. In cases where NNAPI drivers are missing, all computations in the tests from the first group also fall back on CPU and are using the same instruction set.

The resolution of input images used in the tests was chosen so that all devices with at least 2GB of RAM and the majority of devices with 1GB of RAM should have enough memory to run all tests. The test is considered to be passed when the network was able to successfully process at least one image within the allocated time. In particular, during the internal testing all devices with 1GB of RAM (\eg Samsung Galaxy S2/S3 mini, HTC One X, FiiO X7, etc.) were able to run all models after a fresh restart.

\begin{figure}[t!]
\centering
\resizebox{0.94\linewidth}{!}
{
\includegraphics[width=1.0\linewidth]{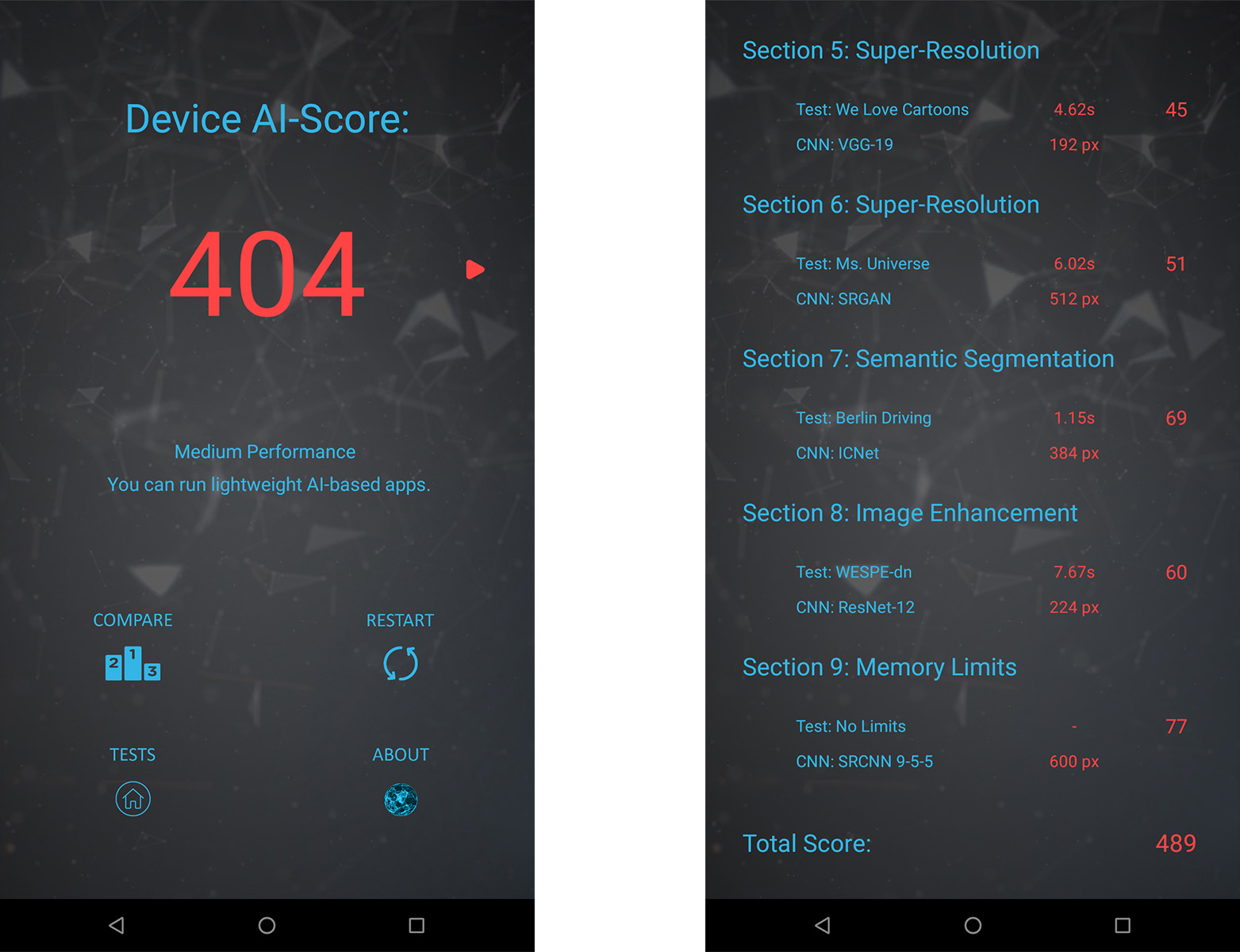}
}
\caption{\small{Benchmark results displayed after the end of the tests.}}
\label{fig:results-vis}
\end{figure}

Each of the first eight tests has a predefined time limit: 25, 40, 40, 30, 40, 50, 20 and 25 seconds, respectively. The last test does not have a time limit~--- images of increasing resolution are processed until the device runs out of memory. The running time for each test is computed as an average over the set of images processed within the specified time.
When more than two images are handled, the processing time for the first two ones is not considered as it might comprise additional time expenses associated with network initialization and memory allocation. The scores for the first eight tests are computed inversely proportional to the corresponding average runtimes; the score for the  memory test is proportionate to the maximum image size that the network was able to process. The final AI score (fig.~\ref{fig:results-vis}) is calculated as a weighted sum of the scores obtained in these nine tests and represents the aggregated AI performance of a particular device. The weight coefficients for these tests were calibrated based on the results obtained on Google Pixel 2 running Android P with disabled NNAPI in all tests.

\begin{table*}[t!]
\centering
\resizebox{2.0\columnwidth}{!}
{
\begin{tabular}{l|ccc|ccccc}
Test & 1 & 2 & 3 & 4 & 5 & 6 & 7 & 8 \\[0.1cm]
\hline\\
Task \, & \, Classification \, & \, Classification \, & \, Face Recognition \, & \, Deblurring \, & \, Super-Resolution \, & \, Super-Resolution \, & \, Segmentation \, & \, Enhancement \, \\
Architecure \, & MobileNet & Inception-V3 & Inc-ResNet-V1 & SRCNN & VGG-19 & SRGAN (ResNet-16) & ICNet & DPED (ResNet-4) \\
Resolution, px \, & \, 224$\times$224 \, & \, 346$\times$346 \, & \, 512$\times$512 \, & \, 300$\times$300 \, & \, 192$\times$192 \, & \, 512$\times$512 \, & \, 384$\times$576 \, & \, 128$\times$192 \\
Parameters \, & \, 4.2M \, & \, 27.1M \, & \, 22.8M \, & \, 69K \, & \, 665K \, & \, 1.5M \, & \, 6.7M \, & \, 400K \\
Size, MB \, & \, 4.3 \, & \, 96 \, & \, 92 \, & \, 0.3 \, & \, 2.7 \, & \, 6.2 \, & \, 27 \, & \, 1.6 \\
Quantized \, & \, yes \, & \, no \, & \, no \, & \, no \, & \, no \, & \, no \, & \, no \, & \, no \\
NNAPI support \, & \, yes \, & \, yes \, & \, no \, & \, yes \, & \, yes \, & \, no \, & \, no \, & \, yes \\
Consumed RAM \, & \, 20MB \, & \, 170MB \, & \, 240MB \, & \, 290MB \, & \, 110MB \, & \, 310MB \, & \, 60MB \, & \, 120MB \\
\end{tabular}
}

\vspace{2.6mm}
\caption{Summarized characteristics of deep learning models used in the AI Benchmark}

\label{networks-summary}
\vspace{0.2cm}
\end{table*}

\section{Benchmark Results}

In this section, we present quantitative benchmark results obtained from over 10,000 mobile devices tested in the wild. The scores of each device/SoC are presented in tables~\ref{ranking-phones} and~\ref{ranking-socs} that are showing average processing time per one image for each test/network, maximum possible image resolution that can be processed by SRCNN model and the total aggregated AI score. The scores were calculated by averaging all obtained results of the corresponding devices/SoCs after removing the outliers. The description of the results is provided below.

\subsection{Neural Networks}

Table~\ref{networks-summary} summarizes the details of all deep learning architectures included in the benchmark. The results in tables~\ref{ranking-phones} and~\ref{ranking-socs} are quite consistent with the theoretical expectations of the relative processing time and memory consumed by the networks. In particular, the quantized MobileNet CNN from the first test requires about 3-4 times less RAM than the same float model, and its speed on CPU is generally an order of magnitude faster compared to Inception-V3 CNN. The third face recognition test is dealing with images with a twice larger area and exhibits around 2x longer inference times than the second one, meaning that the performances of Inception-ResNet-V1 and Inception-V3 are quite comparable. In image-to-image processing tasks, the most efficient model is ICNet since the computations there are mainly done on the downscaled images/feature maps. The same approach is used in the SRGAN model where the original image is downsampled to 128$\times$128 pixels and processed in this resolution till the last two layers that are performing its upscaling to the original size. Therefore, despite using 12 residual blocks, the processing time here still remains reasonable, though the required RAM is quite high due to the downscaling/upscaling layers working with 512$\times$512px images. The DPED network from the image enhancement task contains 4 residual blocks and is processing images without downsampling, therefore the processing time here should be roughly $\frac{128\times128\times12}{128\times192\times4}$ = 2 times faster than in the previous case, as seen in practice. The VGG-19 model from the fifth test is the most resource-consuming among all considered CNNs~--- since it consists of 19 convolutional layers, it should be theoretically around $\frac{19}{12}$ = 1.6 times slower than the DPED network (the size of their convolutional layers is similar), though the RAM consumption should lie in the same range as it is primarily defined by the dimensions of the largest convolutional layer. Finally, the SRCNN model is much faster than both the VGG-19 and DPED networks, and the amount of consumed memory here is also quite similar due to the aforementioned reason. The size of the highest image resolution that can be processed by SRCNN is growing linearly with the amount of total (free) RAM of the device, though due to a bug in NNAPI this does not hold true for devices with Android 8.1+ as they are generally consuming much more RAM. We should also note that all previous conclusions are based on the results from devices not supporting hardware acceleration, since it might significantly alter the results in tests 1, 2, 4, 5, 8 and 9 that can run with NNAPI on dedicated hardware.

\subsection{Smartphones and mobile chipsets}

\begin{table*}[t!]
\centering
\resizebox{2.0\columnwidth}{!}
{
\begin{tabular}{l|ccc|cccccccccc}
Model \, & \, SoC \, & \, RAM \, & \, Android \, & \, Test 1, \, & \, Test 2, \, & \, Test 3, \, & \, Test 4, \, & \, Test 5, \, & \, Test 6, \, & \, Test 7, \, & \, Test 8, \, & \, Test 9, \, & \, AI-Score \\
&  &  &  & ms & ms & ms & ms & ms & ms & ms & ms & 100 px &  \\ [0.1cm]
\hline\\
Huawei P20 Pro \, & \, HiSilicon Kirin 970 \, & \, 6GB \, & \, 8.1 \, & \, 144 \, & \, 130 \, & \, 2634 \, & \, 279 \, & \, 241 \, & \, 4390  \, & \, 779 \, & \, 193 \, & \, 6 \, & \, 6519 \\
OnePlus 6 \, & \, Snapdragon 845/DSP \, & \, 8GB \, & \, 9.0 \, & \, 24 \, & \, 892 \, & \, 1365 \, & \, 928 \, & \, 1999 \, & \, 2885  \, & \, 303 \, & \, 1244 \, & \, 5 \, & \, 2053 \\
HTC U12+ \, & \, Snapdragon 845 \, & \, 6GB \, & \, 8.0 \, & \, 60 \, & \, 620 \, & \, 1433 \, & \, 1229 \, & \, 2792 \, & \, 3542  \, & \, 329 \, & \, 1485 \, & \, 11 \, & \, 1708 \\
Samsung Galaxy S9+ \, & \, Exynos 9810 Octa \, & \, 6GB \, & \, 8.0 \, & \, 148 \, & \, 1208 \, & \, 1572 \, & \, 958 \, & \, 1672 \, & \, 2430  \, & \, 612 \, & \, 1230 \, & \, 8 \, & \, 1628 \\
Samsung Galaxy S8 \, & \, Exynos 8895 Octa \, & \, 4GB \, & \, 8.0 \, & \, 134 \, & \, 731 \, & \, 1512 \, & \, 1197 \, & \, 2519 \, & \, 3039  \, & \, 428 \, & \, 1422 \, & \, 6 \, & \, 1413 \\
Motorola Z2 Force \, & \, Snapdragon 835 \, & \, 6GB \, & \, 8.0 \, & \, 85 \, & \, 823 \, & \, 1894 \, & \, 1513 \, & \, 3568 \, & \, 4302  \, & \, 381 \, & \, 1944 \, & \, 11 \, & \,  1384 \\
OnePlus 3T \, & \, Snapdragon 821 \, & \, 6GB \, & \, 8.0 \, & \, 106 \, & \, 776 \, & \, 1937 \, & \, 1707 \, & \, 3624 \, & \, 4427  \, & \, 365 \, & \, 1982 \, & \, 10 \, & \,  1302 \\
Lenovo ZUK Z2 Pro \, & \, Snapdragon 820 \, & \, 6GB \, & \, 8.0 \, & \, 115 \, & \, 909 \, & \, 2099 \, & \, 1747 \, & \, 3683 \, & \, 4363  \, & \, 313 \, & \, 2030 \, & \, 11 \, & \,  1300 \\
Google Pixel 2 \, & \, Snapdragon 835 \, & \, 4GB \, & \, 9.0 \, & \, 143 \, & \, 1264 \, & \, 1953 \, & \, 1168 \, & \, 2104 \, & \, 4219  \, & \, 394 \, & \, 1360 \, & \, 4 \, & \,  1293 \\
Google Pixel \, & \, Snapdragon 821 \, & \, 4GB \, & \, 9.0 \, & \, 116 \, & \, 867 \, & \, 1838 \, & \, 1287 \, & \, 2489 \, & \, 4125  \, & \, 365 \, & \, 1568 \, & \, 4 \, & \,  1260 \\
Nokia 7 plus \, & \, Snapdragon 660 \, & \, 4GB \, & \, 9.0 \, & \, 136 \, & \, 944 \, & \, 2132 \, & \, 1320 \, & \, 2519 \, & \, 4641  \, & \, 475 \, & \, 1509 \, & \, 5 \, & \,  1183 \\
Asus Zenfone 5 \, & \, Snapdragon 636 \, & \, 4GB \, & \, 8.0 \, & \, 110 \, & \, 1055 \, & \, 2405 \, & \, 1910 \, & \, 4271 \, & \, 4877  \, & \, 515 \, & \, 2330 \, & \, 7 \, & \,  1028 \\
Google Pixel C \, & \, Nvidia Tegra X1 \, & \, 3GB \, & \, 8.0 \, & \, 105 \, & \, 1064 \, & \, 2585 \, & \, 2104 \, & \, 4546 \, & \, 5036  \, & \, 429 \, & \, 2439 \, & \, 6 \, & \,  980 \\
Huawei Honor 8 Pro \, & \, HiSilicon Kirin 960 \, & \, 6GB \, & \, 8.0 \, & \, 121 \, & \, 1720 \, & \, 3163 \, & \, 1943 \, & \, 4791 \, & \, 5719  \, & \, 1082 \, & \, 2764 \, & \, 9 \, & \,  917 \\
Sony XA2 Ultra \, & \, Snapdragon 630 \, & \, 4GB \, & \, 8.0 \, & \, 170 \, & \, 1653 \, & \, 3424 \, & \, 2638 \, & \, 5497 \, & \, 6338  \, & \, 685 \, & \, 3166 \, & \, 9 \, & \,  799 \\
Meizu Pro 7 Plus \, & \, Mediatek Helio X30 \, & \, 6GB \, & \, 7.0 \, & \, 327 \, & \, 3357 \, & \, 4550 \, & \, 2215 \, & \, 4971 \, & \, 5502  \, & \, 1666 \, & \, 2651 \, & \, 10 \, & \,  785 \\
BlackBerry Keyone \, & \, Snapdragon 625 \, & \, 4GB \, & \, 7.1 \, & \, 160 \, & \, 1695 \, & \, 3525 \, & \, 2780 \, & \, 6150 \, & \, 7164  \, & \, 780 \, & \, 3628 \, & \, 9 \, & \,  776 \\
Sony X Compact \, & \, Snapdragon 650 \, & \, 3GB \, & \, 8.0 \, & \, 111 \, & \, 1804 \, & \, 3566 \, & \, 2469 \, & \, 5789 \, & \, 6846  \, & \, 835 \, & \, 3527 \, & \, 6 \, & \,  738 \\
Xiaomi Redmi 5 \, & \, Snapdragon 450 \, & \, 3GB \, & \, 7.1 \, & \, 188 \, & \, 1753 \, & \, 3707 \, & \, 3020 \, & \, 6144 \, & \, 7144  \, & \, 751 \, & \, 3580 \, & \, 8 \, & \, 706 \\
Huawei Nexus 6P \, & \, Snapdragon 810 \, & \, 3GB \, & \, 8.0 \, & \, 106 \, & \, 1962 \, & \, 4113 \, & \, 3389 \, & \, 8155 \, & \, 9805  \, & \, 930 \, & \, 4733 \, & \, 7 \, & \, 658 \\
Meizu MX6 \, & \, Mediatek Helio X20 \, & \, 4GB \, & \, 7.1 \, & \, 183 \, & \, 2217 \, & \, 4981 \, & \, 3906 \, & \, 9245 \, & \, 10551  \, & \, 936 \, & \, 4870 \, & \, 9 \, & \, 641 \\
HTC U Play \, & \, Mediatek Helio P10 \, & \, 3GB \, & \, 6.0 \, & \, 239 \, & \, 2061 \, & \, 4303 \, & \, 3563 \, & \, 7537 \, & \, 10116  \, & \, 989 \, & \, 4368 \, & \, 7 \, & \, 561 \\
Xiaomi Redmi 4X \, & \, Snapdragon 435 \, & \, 3GB \, & \, 7.1 \, & \, 246 \, & \, 2640 \, & \, 5428 \, & \, 4155 \, & \, 8575 \, & \, 9979  \, & \, 1229 \, & \, 5030 \, & \, 8 \, & \, 537 \\
Samsung Galaxy J7 \, & \, Exynos 7870 Octa \, & \, 3GB \, & \, 7.0 \, & \, 278 \, & \, 2092 \, & \, 4648 \, & \, 3881 \, & \, 8495 \, & \, 9644  \, & \, 941 \, & \, 4699 \, & \, 3 \, & \, 455 \\
LG Nexus 5 \, & \, Snapdragon 800 \, & \, 2GB \, & \, 4.4 \, & \, 332 \, & \, 2182 \, & \, 5080 \, & \, 5732 \, & \, 9625 \, & \, 12375  \, & \, 1299 \, & \, 5948 \, & \, 3 \, & \, 387 \\
Asus Zenfone 2 \, & \, Intel Atom Z3580 \, & \, 2GB \, & \, 5.0 \, & \, 1507 \, & \, 2433 \, & \, 6188 \, & \, 4337 \, & \, 12878 \, & \, 15128  \, & \, 1176 \, & \, 6947 \, & \, 3 \, & \, 318 \\
Motorola Moto C \, & \, Mediatek MT6737 \, & \, 1GB \, & \, 7.0 \, & \, 414 \, & \, 3394 \, & \, 7761 \, & \, 6356 \, & \, 14760 \, & \, 16721  \, & \, 1668 \, & \, 7856 \, & \, 3 \, & \, 283 \\
Samsung Galaxy S3 \, & \, Exynos 4412 Quad \, & \, 1GB \, & \, 4.3 \, & \, 553 \, & \, 4640 \, & \, 10321 \, & \, 7587 \, & \, 17187 \, & \, 21904  \, & \, 2059 \, & \, 9291 \, & \, 2 \, & \, 216 \\
Fly Nimbus 15 \, & \, Spreadtrum SC9832 \, & \, 1GB \, & \, 7.0 \, & \, 538 \, & \, 5103 \, & \, 12618 \, & \, 7594 \, & \, 19174 \, & \, 22758  \, & \, 2094 \, & \, 9935 \, & \, 2 \, & \, 202 \\
Huawei Ascend P1 \, & \, TI OMAP 4460 \, & \, 1GB \, & \, 4.1 \, & \, 482 \, & \, 7613 \, & \, 25105 \, & \, 12667 \, & \, 30743 \, & \, 35417  \, & \, 4015 \, & \, 18836 \, & \, 2 \, & \, 140 \\
\end{tabular}
}
\vspace{2.6mm}
\caption{Benchmark results for several Android devices, a full list is available at:\, \small{\url{http://ai-benchmark.com/ranking}}}

\label{ranking-phones}
\vspace{0.2cm}
\end{table*}

The results in tables~\ref{ranking-phones} and~\ref{ranking-socs} show the performance of several selected Android smartphones and chipsets obtained with the AI Benchmark; the actual full list is available on the project website: \url{http://ai-benchmark.com}. Before going into details, we would first like to mention several Android NNAPI bugs that are currently affecting some results presented in the tables. First of all, due to a bug in Android 8.1 with default NNAPI drivers, the performance of (convolutional) operations is twice as slow as when these drivers are disabled. Therefore, when calculating the average runtime for different SoCs presented in table~\ref{ranking-socs}, we omitted the results from the phones with this issue. While Huawei phones with Android 8.1 and the Kirin 970 chipset were using their own customized NNAPI implementation, it still suffered from a different bug~--- after a long standby the clock speed of Kirin's NPU drops and does not return back until the phone is rebooted. The results in both tables represent the scores obtained from Huawei devices that were recently restarted. Finally, the RAM consumption on devices using Android NNAPI might be up to 2$\times$ higher in image-to-image processing tests due to the ByteBuffer issue described in Section 3.2; its consequences can be observed in the last memory test.

Below we summarize the results for each SoC manufacturer and describe the performance of the corresponding chipsets present on the market.

\textbf{$\bullet$ Qualcomm.} Snapdragon chipsets can now provide hardware acceleration for quantized neural networks (when Qualcomm's NNAPI drivers are present), while float models are not yet supported by existing commercial devices. The first smartphone to contain these drivers is the OnePlus 6 with Snapdragon 845 SoC and the latest Android P firmware. It can run the quantized MobileNet model under 25ms on the Hexagon DSP which is considerably faster than the corresponding CPU speed (60-65ms). A similar performance can be expected from Snapdragon 670/710 chipsets containing the same Hexagon 685 DSP; Snapdragon 835 with Hexagon 682 and Snapdragon 636/660/820/821 with Hexagon 680 from the same Qualcomm 68x DSP family should come with a somewhat longer runtime.

While there exist no official tests of Qualcomm's NNAPI drivers supporting acceleration for float models, the Snapdragon 625 SoC, with (presumably) a beta version of these drivers using the integrated Adreno 506 GPU, can provide up to 2x speed-up compared to a CPU-based execution. While the performance of Adreno 506 is around 130 GFLOPs, this means that Adreno 630 (727 GFLOPs) present in Snapdragon 845 SoC can potentially provide a speed-up by a factor of 3-4, though the exact number might vary a lot.

As to CPU performance measured in relation to matrix/deep learning computations, currently the most powerful Qualcomm core is the Kryo 385 Gold present in the Snapdragon 845 SoC. It exhibits around a 30\% improvement over the Kryo 280 cores from Snapdragon 835. Interestingly, the latter ones demonstrate a similar or slightly degraded performance (per GHz) compared to the first Kryo generation in the Snapdragon 820 SoC with a custom non-Cortex based design, that despite having only 4 cores is still slightly faster than the Snapdragon 636/660 with newer Kryo 260 cores. The previous Krait microarchitecture represented by the Snapdragon 800/801 from 2013 is still showing competitive results, outperforming the majority of SoCs from the 2xx, 4xx and 6xx families or even subsequently presented 810 and 808 chipsets based on the Cortex-A57 microarchitecture. We also note that customized Qualcomm CPU cores are generally showing a better performance than the default Arm Cortex architectures.

\textbf{$\bullet$ Huawei.} Though the CPU performance of HiSilicon SoCs is not as impressive as in Qualcomm's case, its NPU integrated into the Kirin 970 provides a dramatic speed-up for float deep learning models. In particular, depending on the task it demonstrates 7-21 times faster inference compared to its CPU and 4-7 times better performance compared to the overall best CPU results. In tests 2, 4, 5, 8 that are supporting hardware acceleration, it requires on average 132, 274, 240 and 193 milliseconds to process one image, respectively. The only main weakness of this NPU is the lack of acceleration support for quantized models~--- in the first test all computations are running on CPU with an average processing time of 160ms per image, which is significantly higher than the corresponding results of the Snapdragon 845 with enabled DSP. Though this problem can be solved by implementing a quantized mode in Kirin's NNAPI drivers, at the present time this functionality is still under development.

Regarding other HiSilicon chipsets, they are now not providing acceleration for AI apps, and thus all computations are running on CPUs only. Since all HiSilicon's SoCs are based on standard Arm Cortex cores, their performance is also quite similar to other chipsets with the same Cortex architectures.

\begin{table*}[t!]
\centering
\resizebox{2.0\columnwidth}{!}
{
\begin{tabular}{l|c|cccccccc}
SoC \, & \, Cores \, & \, Test 1, \, & \, Test 2, \, & \, Test 3, \, & \, Test 4, \, & \, Test 5, \, & \, Test 6, \, & \, Test 7, \, & \, Test 8, \\
&  & ms & ms & ms & ms & ms & ms & ms & ms \\ [0.1cm]
\hline\\
HiSilicon Kirin 970 \, & \, CPU (4x2.4 GHz A73 \& 4x1.8 GHz A53) + NPU \, & \, 160 \, & \, 132 \, & \, 2586 \, & \, 274 \, & \, 240   \, & \, 4848  \, & \, 742 \, & \, 193 \\
Mediatek Helio P60 Dev \, & \, CPU (4x A73 + 4x A53) + GPU (Mali-G72 MP3) + APU \, & \, 21 \, & \, 439 \, & \, 2230 \, & \, 846 \, & \, 1419   \, & \, 4499  \, & \, 394 \, & \, 1562 \\
Exynos 9810 Octa \, & \, 8 (4x2.7 GHz Mongoose M3 \& 4x1.8 GHz Cortex-A55) \, & \, 149 \, & \, 1247 \, & \, 1580 \, & \, 956 \, & \, 1661   \, & \, 2450  \, & \, 613 \, & \, 1230 \\
Snapdragon 845 \, & \, 8 (4x2.8GHz Kryo 385 Gold \& 4x1.8GHz Kryo 385 Silver) \, & \, 65 \, & \, 661 \, & \, 1547 \, & \, 1384 \, & \, 3108   \, & \, 3744  \, & \, 362 \, & \, 1756 \\
Exynos 8895 Octa \, & \, 8 (4x2.3 GHz Mongoose M2 \& 4x1.7 GHz Cortex-A53) \, & \, 135 \, & \, 742 \, & \, 1548 \, & \, 1213 \, & \, 2576   \, & \, 3181  \, & \, 451 \, & \, 1492 \\
Snapdragon 835 \, & \, 8 (4x2.45 GHz Kryo 280 \& 4x1.9 GHz Kryo 280) \, & \, 97 \, & \, 855 \, & \, 2027 \, & \, 1648 \, & \, 3771   \, & \, 4375  \, & \, 439 \, & \, 2046 \\
Snapdragon 820 \, & \, 4 (2x2.15 GHz Kryo \& 2x1.6 GHz Kryo) \, & \, 119 \, & \, 839 \, & \, 2074 \, & \, 1804 \, & \, 4015 \, & \, 5055  \, & \, 410 \, & \, 2128 \\
Nvidia Tegra X1 \, & \, 4 (4x1.9 GHz Maxwell) \, & \, 102 \, & \, 925 \, & \, 2328 \, & \, 1811 \, & \, 3824 \, & \, 4437  \, & \, 384 \, & \, 2161 \\
Snapdragon 660 \, & \, 8 (4x2.2 GHz Kryo 260 \& 4x1.8 GHz Kryo 260) \, & \, 115 \, & \, 1025 \, & \, 2299 \, & \, 1806 \, & \, 4072 \, & \, 4695  \, & \, 547 \, & \, 2225 \\
Snapdragon 636 \, & \, 8 (8x1.8 GHz Kryo 260) \, & \, 110 \, & \, 1055 \, & \, 2405 \, & \, 1910 \, & \, 4271 \, & \, 4877  \, & \, 515 \, & \, 2330 \\
Exynos 8890 Octa \, & \, 8 (4x2.3 GHz Mongoose \& 4x1.6 GHz Cortex-A53) \, & \, 139 \, & \, 1810 \, & \, 3314 \, & \, 1536 \, & \, 3594 \, & \, 4717  \, & \, 937 \, & \, 2148 \\
HiSilicon Kirin 955 \, & \, 8 (4x2.5 GHz Cortex-A72 \& 4x1.8 GHz Cortex A53) \, & \, 136 \, & \, 1383 \, & \, 2932 \, & \, 2143 \, & \, 5132 \, & \, 6202  \, & \, 751 \, & \, 2731 \\
\end{tabular}
}
\vspace{2.6mm}
\caption{Benchmark results for several SoCs, the full list available at:\, \small{\url{http://ai-benchmark.com/ranking_processors}}}
\label{ranking-socs}
\vspace{0.2cm}
\end{table*}

\textbf{$\bullet$ MediaTek.} The Helio P60 is the first chipset to get NNAPI drivers for accelerating both float and quantized models. Quantized networks are running on its integrated APU that is showing a performance similar to that of the Hexagon 685 DSP~--- 21ms for processing one image in the first test. Float networks are executed on the Mali-G72 MP3 GPU that provides about 2-5 times acceleration compared to its CPU and 1.5-2x faster runtime than the overall best CPU results. We should mention that all these numbers were obtained on MediaTek's developer phones, while the only Helio P60-based actual device having NNAPI drivers (Vivo V11) is showing slightly worse results.

Other MediaTek chipsets are currently not supporting acceleration for AI applications. They run on CPU cores with standard Arm Cortex designs.

\textbf{$\bullet$ Samsung.} At the time of writing, neither of Samsung's SoCs can provide any acceleration for third-party AI apps: all devices with these chipsets are using default NNAPI drivers. Since the latest Exynos 9810 SoC has the same Mali-G72 graphics as in the MediaTek P60 chipset (but with 12 instead of 3 cores), we can expect an additional speed-up factor of 3-4 for float neural networks if the Arm NN library was integrated by Samsung into its NNAPI drivers. Since all recent Samsung Exynos processors are using Arm Mali GPUs, the same logic can be applied to them just the same.

Depending on the task, Samsung's Mongoose M3 CPU cores can demonstrate significantly better or worse performance compared to custom Kryo 385 cores in the Snapdragon 845, but their overall performance can be considered quite comparable. The Mongoose M2 microarchitecture shows a significant 50\% boost over the first M1 version, while the performance of the second (M2) and third (M3) generations is rather similar. One notable issue with the latest Exynos 8895 and 9810 SoCs is related to their integrated power management system responsible for adjusting the CPU performance. It is causing very unstable results on the majority of devices: in particular, several subsequent benchmark runs (with an interval of 10 minutes, ``high performance" mode) on the same Galaxy S9 phone demonstrated up to 50\% variation of the total score, while the results obtained from different devices showed an even larger variation (\eg 200-800 ms in the seventh test). Currently, there is no way to have external control over different performance modes as they are selected automatically based on the integrated logic.

\textbf{$\bullet$ Others.} We have obtained results from a number of other chipsets that are either not widely used (\eg Spreadtrum) or deprecated by their manufacturers (\eg Intel Atom, Nvidia Tegra, TI OMAP). Especially interesting in the context of AI and deep learning are Nvidia Tegra platforms that are supporting CUDA~\cite{kirk2007nvidia} and cuDNN~\cite{chetlur2014cudnn} GPU-accelerated libraries of primitives for deep neural networks. Unfortunately, no new devices using Nvidia SoCs were released since 2015, and the existing ones are already deprecated and will not get (NNAPI) drivers for accelerating machine learning mobile frameworks.

\section{Discussion}

Software and hardware support for machine learning on mobile devices is now evolving extremely fast, with various milestone releases announced each several months. While they are certainly bringing new possibilities and higher levels of performance, the current lack of standardized requirements and publicly available specifications does not always allow for an objective assessment of their real advantages and limitations. Below we would like to summarize our experience of working with mobile machine learning frameworks and chipsets providing hardware acceleration via NNAPI drivers.

Currently, the easiest way to start using deep learning on Android is to go for a mature and relatively stable TensorFlow Mobile framework. It was introduced more than two years ago, and all major issues are already solved, while plenty of information on smaller problems is available on various specialized websites. If hardware acceleration is one of the critical problems, TensorFlow Lite can still be an option, but we would not recommend using it now for anything more complicated than image classification with MobileNet or Inception CNNs as there still might be occasional problems with non-standard network architectures on some mobile platforms. We can also mention that migrating from TF Mobile to Lite is relatively easy since they are using very similar Android programming interfaces (the biggest difference will be in converting pre-trained models to .tflite instead of .pb format), and thus can be done later when TF Lite gets better support. If the application is targeted at some specific device or SoC, the corresponding proprietary SDK can also be used, though in this case the development might not be so easy and convenient. Regarding Caffe2 Mobile and other less widespread frameworks, their communities are now very small, which means that almost no tutorials and problem descriptions are available on the internet, thus all appearing problems might be primarily solved only by creating new issues in the corresponding github repositories.

Hardware acceleration for AI algorithms on Android devices is now an even more controversial topic. At the time of writing, the fastest runtime for conventional float neural networks is shown by Huawei devices with Kirin 970 chipsets that at the time of their presentation were significantly ahead of the market. Yet, we prefer to stay neutral regarding the future perspectives, as our analysis has demonstrated that almost all SoC manufacturers have the potential to achieve similar results in their new chipsets. The real situation will become clear at the beginning of the next year when the first devices with the Kirin 980, the MediaTek P80 and the next Qualcomm and Samsung Exynos premium SoCs will appear on the market. Besides the performance, we would also like to look at their power efficiency since a significant battery drain might restrict their usage to a few standard in-camera processing techniques.

The last topic that we want to address here is the use of quantized networks. Their current applicability is rather limited, as there are still no standard and reliable tools for quantizing networks trained even for image classification, not to mention more complex tasks. At the moment we can expect two different ways of development in this area. In the first case, the problem of quantization will be largely solved at some point, and the majority of neural networks deployed on smartphones will be quantized. In the second case, specific NPUs supporting float networks will become even more powerful and efficient, and the need for quantization will disappear as this happened to many optimized solutions developed due to the lack of computational power in the past. Since we cannot easily predict the future outcome, we will still be using a mixture of quantized and float models in the benchmark with predominance of the second ones, though in the future releases the corresponding ratio might be significantly altered.

Since currently there are still many important open questions that might be answered only with new major software and hardware releases related to machine learning frameworks and new dedicated chipsets, we are planning to publish regular benchmark reports describing the actual state of AI acceleration on mobile devices, as well as changes in the machine learning field and the corresponding adjustments made in the benchmark to reflect them. The latest results obtained with the AI Benchmark and the description of the actual tests will also be updated monthly on the project website: \url{http://ai-benchmark.com}. Additionally, in case of any technical problems or some additional questions you can always contact the first two authors of this paper.

\section{Conclusions}

In this paper, we discussed the latest achievements in the area of machine learning and AI in the Android ecosystem. First, we presented an overview of all currently existing mobile chipsets that can be potentially used for accelerating the execution of neural networks on smartphones and other portable devices, and described popular mobile frameworks for running AI algorithms on mobile devices. We presented the AI Benchmark that measures different performance aspects associated with running deep neural networks on smartphones and other Android devices, and discussed the real-world results obtained with this benchmark from over 10,000 mobile devices and more than 50 different mobile SoCs. Finally, we discussed future perspectives of software and hardware development related to this area and gave our recommendations regarding the current deployment of deep learning models on Android devices.

{\footnotesize
\bibliographystyle{splncs}

}

\end{document}